\documentclass[utf8]{frontiersSCNS} 

\usepackage{url,hyperref,lineno,microtype,subcaption}
\usepackage[onehalfspacing]{setspace}

\usepackage{multirow}
\usepackage{colortbl}

\def\keyFont{\fontsize{8}{11}\helveticabold }
\def\firstAuthorLast{Allred {et~al.}} 
\def\Authors{Jason M. Allred\,$^{1,*}$ and Kaushik Roy\,$^{1}$}

\def\Address{$^{1}$Nanoelectronics Research Laboratory, Electrical and Computer Engineering Department, Purdue University, West Lafayette, IN, USA}
\def\corrAuthor{Jason M. Allred}

\def\corrEmail{allredj@purdue.edu}

\begin{document}
\onecolumn
\firstpage{1}

\title[Controlled Forgetting]{Controlled Forgetting: Targeted Stimulation and Dopaminergic Plasticity Modulation for Unsupervised Lifelong Learning in Spiking Neural Networks} 

\author[\firstAuthorLast ]{\Authors} 
\address{} 
\correspondance{}

\extraAuth{}

\maketitle

\begin{abstract}

\section{}
Stochastic gradient descent requires that training samples be drawn from a uniformly random distribution of the data.
For a deployed system that must learn online from an uncontrolled and unknown environment, the ordering of input samples often fails to meet this criterion, making lifelong learning a difficult challenge.
We exploit the locality of the unsupervised Spike Timing Dependent Plasticity (STDP) learning rule to target local representations in a Spiking Neural Network (SNN) to adapt to novel information while protecting essential information in the remainder of the SNN from catastrophic forgetting.
In our Controlled Forgetting Networks (CFNs), novel information triggers stimulated firing and heterogeneously modulated plasticity, inspired by biological dopamine signals, to cause rapid and isolated adaptation in the synapses of neurons associated with outlier information.
This targeting controls the forgetting process in a way that reduces the degradation of accuracy for older tasks while learning new tasks.
Our experimental results on the MNIST dataset validate the capability of CFNs to learn successfully over time from an unknown, changing environment, achieving 95.36\% accuracy, which we believe is the best unsupervised accuracy ever achieved by a fixed-size, single-layer SNN on a completely disjoint MNIST dataset.

\tiny
 \keyFont{ \section{Keywords:} lifelong learning, continual learning,
 catastrophic forgetting, controlled forgetting, dopaminergic learning, spiking neural networks, spike timing dependent plasticity, stability-plasticity dilemma}
\end{abstract}

%Word Count: $\approx$ 6,487. Figure Count: 11. Table Count: 1.
%Preprint: \href{https://arxiv.org/abs/1902.03187}{https://arxiv.org/abs/1902.03187}.

\section{Introduction}

Artificial neural networks have enabled computing systems to successfully perform tasks previously out of reach for traditional computing, such as image and audio classification.
These networks, however, are typically trained offline and do not update during deployed inference.
One of the current obstacles preventing fully autonomous, unsupervised learning in dynamic environments while maintaining efficiency is the \textit{stability-plasticity dilemma}, or the challenge of ensuring that the system can continue to quickly and successfully learn from and adapt to its current environment while simultaneously retaining and applying essential knowledge from previous environments (\cite{GROSSBERG198723}).

\begin{figure}
\begin{center}
\includegraphics[width=85mm]{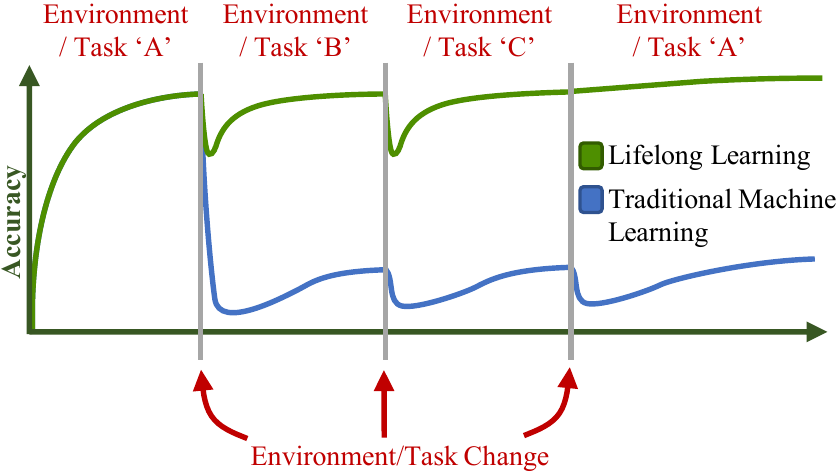}
\end{center}
\caption{The goal of lifelong learning: quickly and correctly adapt to new environments while retaining essential information from previous environments.}
\label{fig_lifelong_learning}
\end{figure}

There have been a handful of terms used in literature to describe the process of learning from data that is temporally distributed inhomogeneously, such as the terms incremental learning, sequential learning, continual learning, and lifelong learning.
In this work, we will use the term ``lifelong learning."
\textit{Lifelong learning} is the process of successfully learning from new data while retaining useful knowledge from previously encountered data that is statistically different, often with the goal of sequentially learning differing tasks while retaining the capability to perform previously learned tasks without requiring retraining on data for older tasks (see Figure \ref{fig_lifelong_learning}).
When traditional artificial neural networks are presented with changing data distributions, more rigid parameters interfere with adaption, while more flexibility causes the system to fail to retain important older information, a problem called \textit{catastrophic interference} or \textit{catastrophic forgetting}.
Biological neuronal systems don’t seem to suffer from this dilemma.
We take inspiration from the brain to help overcome this obstacle.

To avoid catastrophic forgetting, important information from older data must be protected while new information is learned from novel data.
Non-local learning rules may not provide such isolation.
Localized learning, on the other hand, may provide the desired segmentation while also being able to perform unsupervised learning, which is critical for lifelong learning in unknown environments.
Spike Timing Dependent Plasticity (STDP) is a localized biological Hebbian learning process where a synaptic weight's adjustment is a function of the timing of the \textit{spikes}, or firing events, of its locally connected pre- and post-synaptic neurons.
Spiking Neural Networks (SNNs), which have been explored for their potential energy advantages due to sparse computing (\cite{han2017deepspikingenergy}), have been shown to perform successful unsupervised clustering tasks with STDP (\cite{Diehl_MNIST}).

However, even though STDP learning is localized, it is still susceptible to catastrophic forgetting because the algorithms that employ STDP are traditionally designed for randomized input ordering.
Certain features, such as homeostasis, attempt to distribute the effect of input groupings globally in order to benefit from the full network.
Without a temporally uniform distribution of classes, traditional STDP algorithms still lose important older information, which is either replaced by or corrupted with information from newer samples (\cite{allred2016forcedfiring}).

We present a new learning paradigm, inspired by the dopamine signals in mammalian brains that non-uniformly modulate synaptic plasticity.
We create Controlled Forgetting Networks (CFNs) that address the stability-plasticity dilemma with rapid/local learning from new information, rather than the traditional gradual/global approach to learning.
Our approach allows fixed-size CFNs to successfully perform unsupervised learning of sequentially presented tasks without catastrophically forgetting older tasks.

Many recent papers have tackled the challenge of lifelong learning without catastrophic forgetting, but they are not designed to target the goal of this paper, which is autonomous learning on a deployed neuromorphic system.
This goal requires real-time unsupervised learning, energy efficiency, and fixed network resources.
\cite{lee2017incremental}, \cite{rusu2016progressive}, \cite{srivastava2013compete}, \cite{fern2017pathnet}, \cite{li2018withoutforgetting}, \cite{kirkpatrick2017overcoming}, \cite{2019arXiv190409330B}, \cite{wang2015evolving}, \cite{wang2014adaptive}, \cite{wysoski2006adaptation}, \cite{du2019singlenet}, and \cite{2017arXiv171109601A} all employ supervised or reinforcement learning methods, in some way provide the network with the knowledge of when a task change occurs, or provide access to previous samples for retraining.
For example, the work by \cite{2017arXiv171109601A} requires that the system be allowed a parameter-``importance update" period on the older task(s) before proceeding to a new task.
Similarly, \cite{panda2018asp} requires that samples from earlier distributions be presented in disproportionately larger quantities than later distributions to avoid catastrophic forgetting, which would require knowledge of a task change.
Additionally, \cite{lee2017incremental}, \cite{rusu2016progressive}, \cite{srivastava2013compete}, \cite{fern2017pathnet}, \cite{li2018withoutforgetting}, \cite{kirkpatrick2017overcoming}, and \cite{am2018closedloop} are also not applicable to localized learning rules that may be employed on spiking networks.
And \cite{wysoski2006adaptation}, \cite{wang2017adaptive}, and \cite{dhoble2012evolving} are morphological systems that do not work with static-sized networks, which would exclude them from direct mapping onto physical hardware implementations.

\section{Materials and Methods}

\subsection{The Challenge of Lifelong Learning}

Backpropagation has proven a successful learning algorithm for deep neural networks.
The accuracy of this approach depends on proper stochastic gradient descent or SGD, also known as incremental gradient descent, in which many small, global adjustments to network weights are performed while iterating over samples from a training dataset.
These samples, however, must be drawn from a random distribution of the dataset—hence the name ``stochastic" gradient descent—intermixing the classes so that each class can affect the direction of descent for correct error minimization throughout the entire training process.

The need to draw training samples from a random distribution is an obstacle for on-line learning, especially when the system encounters novel data.
Backpropagation in an on-line system for real-time learning proves difficult when the input from the environment is uncontrolled and unknown.
With traditional SGD, the system typically has three choices to attempt learning from novel data: (1) train normally on inputs in the order seen; (2) periodically go offline and retrain from an updated dataset; (3) maintain an online storage of previous samples to intermix with the new samples, providing a simulated random sampling.
The latter two choices are costly and inhibit real-time learning, while the first catastrophically violates SGD.

\subsubsection{Catastrophic Forgetting due to Global Interference}

If a uniformly randomized order is not provided, e.g. samples are grouped by class and classes are presented sequentially to the network, then the gradient descent followed by latter samples will likely disagree with the direction from previous samples.    
This conflict causes the network to fail to reach an     error minimum that respects older tasks, as at each period of time in the training process the network essentially attempts to globally optimize for only the current tasks, agnostic as to whether or not that particular direction increases the error for older tasks.
Latter samples erase or corrupt the information learned from previous samples, causing catastrophic forgetting.

One of the largest underlying causes of catastrophic forgetting in backpropagation algorithms is the reliance on a global error.
Calculating weight updates from the current sample's global error means that the current sample may globally affect network weights.
Biological neuronal learning, on the other hand, appears to be significantly localized, with synaptic weight updates being a function of local activity, causing different regions to be responsible for different tasks.
While distributed representations promote generalization in neural networks, rapid learning of novel information may not require significant modifications to low-level distributed representations in a sufficiently trained network.
It has been shown that the IT cortex contains a large-scale spatial organization, or ``shape map," that remains significantly stable over time (\cite{OpdeBeeck2007IT}), even while learning novel information.
It is therefore theorized that the stable lower levels of the visual cortex may be capable of representing the generic structure and composition of never-before-seen inputs with an already-learned understanding of the physical world that remains constant through the remainder of life--for example, an understanding of lines, edges, curves, and colors at the lowest levels and an understanding of rotations, shading, and physical properties at subsequent levels.
Thus, lifelong learning may only need to occur in the last few layers of a neural network, where local learning may sufficiently classify from a read-out of the higher-dimensional generalizations that have been learned previously.

\subsubsection{Catastrophic Forgetting in Localized Learning Due to Homeostasis}

Many leading STDP-trained SNNs employ adaptive thresholding, in which a neuron's firing threshold increases each time it fires and otherwise decays, preventing specific neurons from dominating the receptive field.
Adaptive thresholding helps achieve homeostasis by distributing the firing activity between neurons.
However, adaptive thresholding assumes a temporally random distribution of input samples and often causes catastrophic interference when the environment changes (\cite{allred2016forcedfiring}).
For lifelong learning, adaptive thresholding must be modified to account for long-term variations in spiking activity that would occur when processing temporally variant input distributions.

\subsubsection{The Need for Forgetting}

For successful lifelong learning, there must be network resources available to learn new information.
In a deployed system with finite resources, some forgetting of older knowledge is required to make room for information from new data.
As mentioned earlier, there are morphological systems that logically grow the network to accommodate new information, even employing pruning techniques when necessary if the network grows too large.
However, for our goal of deployed learning on neuromorphic hardware, inserting and removing physical components of the network is not an option, and existing network components must be re-purposed to learn a new task, causing forgetting.

Additionally, in some cases, forgetting may actually be beneficial.
Forgetting outlier data can improve generalizations, and forgetting stale data can allow the system to adapt to a changing environment if new information directly contradicts older information.
Because some forgetting must occur, we seek to control the forgetting process to protect the most vital information, minimizing accuracy loss.

\subsection{Controlled Forgetting with Dopaminergic Learning}
\label{Section_Dop}

\begin{figure}
\begin{center}
\includegraphics[width=85mm]{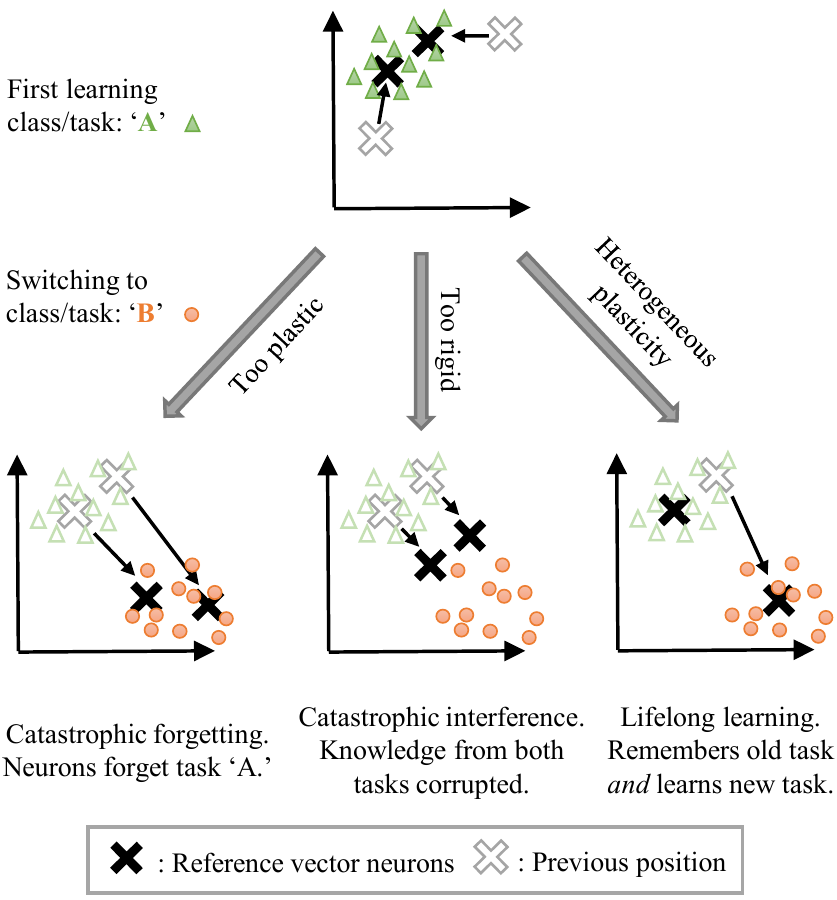}
\end{center}
\caption{The stability-plasticity dilemma in unsupervised clustering. Lifelong learning is achieved with a strategic heterogeneous modulation of synaptic plasticity.}
\label{fig_stab_plast}
\end{figure}

The stability-plasticity dilemma can be addressed by allowing for dynamic, heterogeneously modulated plasticity.
Consider the example of unsupervised clustering where neurons are trained to center on input clusters (see Figure \ref{fig_stab_plast}). 
Temporarily making the synaptic weights of some neurons more plastic while keeping the weights of other neurons more rigid can allow for isolated adaptation by the plastic parameters while protecting the information associated with the rigid parameters.
The challenge then becomes how to dynamically control the plasticity and for which parameters.

STDP embeds local, generalized representations of correlated inputs within the synaptic weights of individual neurons.
Lateral inhibition between neurons, similar to the architecture in \cite{Diehl_MNIST}, creates competition that prevents multiple neurons from learning the same information.
We seek to control the forgetting process by harnessing the segmentation of localized and distinct representations that are created by STDP with competition.
Interference from novel information may be isolated by stimulating specific network elements to adapt to that information, protecting the remainder of the network from change.
The forgetting cause by this interference may be minimized and controlled by targeting network elements associated with less useful information.
We draw on inspiration from biology to heterogeneously modulate STDP learning to perform such isolated adaptation, creating Controlled Forgetting Networks (CFNs).

\subsubsection{Biologically Inspired Dopaminergic Plasticity Modulation}

Dopamine acts as a neuromodulator which gates synaptic plasticity.
Dopamine signals are most commonly thought of as reward signals.
In addition, though, dopamine releases are also associated with encountering novel data, which allows the brain to quickly adapt to new information (\cite{fremaux2016neuromodulation}).
We adopt this concept of novelty-induced plasticity modulation for our goal of local, rapid adaptation.

\begin{figure}
\begin{center}
\includegraphics[width=85mm]{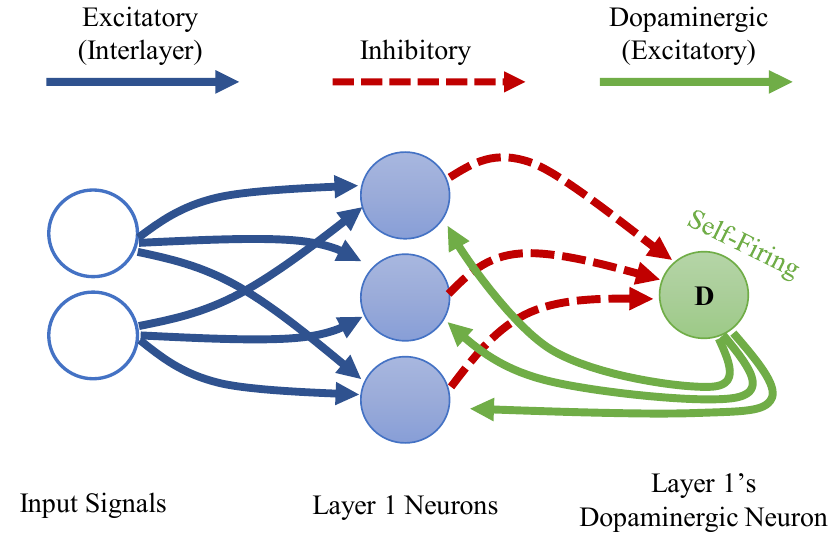}
\end{center}
\caption{Single-layer CFN architecture. The dopaminergic neuron fires when the other neurons on its layer are \textit{not} firing, often a sign of novel information. The firing of the dopaminergic neuron stimulates firing in the other neurons while enhancing plasticity. The dopaminergic weights are trained for heterogeneous stimulation. The other neurons within a layer each have additional laterally inhibitory connections for competition (not shown here). For deeper networks, a separate dopaminergic neuron would be present at each layer.}
\label{fig_dop_arch}
\end{figure}

We mimic a dopamine release by including a \textit{dopaminergic neuron} at each layer (see Figure \ref{fig_dop_arch}) of a CFN.
When an input sample results in little-to-no firing activity at a given layer of neurons, we may assume that it contains information novel to that layer.
We design the dopaminergic neuron with a resting potential higher than its firing potential, giving it a self-firing property.
It is additionally suppressed via inhibitory connections from the other neurons in its layer so that it only spikes when they do not.
This setup allows the dopaminergic neuron to fire only when novel information is detected.

When it fires, the dopaminergic neuron enhances plasticity by temporarily boosting the learning rate of the other neurons in its layer while simultaneously stimulating firing in those other neurons via excitatory synaptic connections that we are calling \textit{dopaminergic weights}.
Because of the lateral inhibition discussed previously, once one of the stimulated neurons fires, it prevents or reduces the probability of the other neighboring neurons from firing.
A neuron with a boosted learning rate then resets its learning rate the next time it fires or receives an inhibitory signal from a neighboring neuron, indicating that one of its neighbors has fired.
Thus, while the dopamine signal is sent to many neurons, only the first neuron(s) to fire undergo the enhanced plasticity, allowing the dopamine signal to perform an isolated targeting for local, rapid adaptation rather than global interference.

Due to the rapid learning that occurs in the presence of dopamine and the lack of traditional homeostatic threshold dynamics, we modify the STDP learning rule for improved stability, discussed later in more detail in Section \ref{subsub_Modified_STDP}.

\subsubsection{Targeted Stimulation for Controlled Forgetting via Trained Dopaminergic Weights}
\label{subsub_dop_weights}

We have addressed how to make the forgetting process local in order to reduce catastrophic interference caused by novel information.
We now address how to control the locality of the forgetting so as to maintain high accuracy for previous tasks.
To minimize accuracy degradation caused by forgetting, we would ideally like to forget outlier or stale information rather than commonly-used or recent information that may be essential for returning to previous tasks, applying knowledge from old tasks to new tasks, or generalizing novel information recently received from the new task.

As a proxy for this categorization, we target neurons with low overall firing frequency (outlier) or less recent firing activity (stale).
The weight of considering firing age over firing frequency is a tunable parameter that controls how much if any preference should be given to more recent tasks.
For the experiments in this paper, we consider all tasks as equally important no matter how recently seen, so we target neurons with low firing frequency.

The dopaminergic weights control how much the dopaminergic neuron stimulates firing activity, and properly training these weights controls the forgetting process.
For our purposes, the dopaminergic weights should be smaller when the post-synaptic neuron it is targeting has a higher firing rate, and vice versa, which is achieved by a simple local learning rule.
The dopaminergic weight depresses each time its post-synaptic neuron fires.
The depression is exponential to maintain positive values.
Otherwise, the dopaminergic weights experience a gradual potentiation to keep them in a manageable range.
The rate of potentiation is irrelevant in our setup as long as it is the same for all dopaminergic weights in the layer because the dopaminergic neuron continues to send the dopaminergic signal until one of the other neurons in the layer fires.
For the experiments in this work, we effect this potentiation by re-normalizing the dopaminergic weights after a depression.

\subsection{Models}

In this subsection, we describe the input, synapse, and neuron models and associated probability distributions that are useful in selecting the appropriate hyperparameters for unsupervised lifelong learning.

\subsubsection{Input Encoding}
\label{subsub_input_encoding}

Input samples are encoded as Poisson spike trains, where the spike rate $\lambda_i$ of an input neuron is proportional to the pixel intensity of input $i$.\footnote{Other input encodings that use time-encoding such as rank-order are beyond the scope of this work and are the subject of future research.}
Thus, the number of spikes in a given time window follows the distribution of a Poisson random variable with an expectation proportional to the input value, while maintaining statistical independence between input spiking events.
Each spike is modeled as a time-shifted delta function.
The precise time of the $k^{th}$ most recent spike from input $i$ is represented as $t_{ik}$.
The time passed since the $k^{th}$ most recent spike from $i$ at time $t$ is represented as $t_{|ik|} = t - t_{ik}$ and follows the distribution of a gamma random variable $T_{|ik|} \sim gamma(\alpha=k;\beta=\lambda_i)$.
The vector of all input rates for each dimension of the given sample is represented as $\vec{\lambda}$.

\subsubsection{Synapse Model}

We model the synaptic connections between neurons as a multiplicative weight which is applied to the delta spike from its pre-synaptic neuron and then added to the membrane potential of its post-synaptic neuron, creating a exponential kernel response.\footnote{Non-instantaneous potentiation kernels, such as the alpha response, are beyond the scope of this work due to the added difficulty to event-driven simulation.} We represent the weight of the synapse connecting input $i$ to neuron $j$ as $w_{ij}$ and the vector of all inputs to neuron $j$ as $\vec{w}_j$.

\subsubsection{Spiking Neuron Model}

We use the common Leaky-Integrate-and-Fire (LIF) neuron model, in which a neuron's membrane potential $v_{mem}$ undergoes a continuous decay according to the differential equation in (\ref{eq_membrane_decay}), where $\tau_{mem}$ is the membrane decay constant and $v_{rest}$ is the resting potential.
The membrane potential is also potentiated or depressed by incoming excitatory or inhibitory signals, respectively.
If the membrane potential reaches or surpasses the neuron's firing threshold $v_{th}$ then the neuron fires, producing an output spike and resetting its potential to $v_{reset}$.
Without loss of generality, we set $v_{rest}$ to zero as a reference voltage.
For model and evaluation simplicity, we also set $v_{reset}$ to zero and have no refractory periods.

\begin{equation} \label{eq_membrane_decay}
\dot{v}_{mem} = \frac{-(v_{mem}-v_{rest})}{\tau_{mem}}
\end{equation}

\subsubsection{Membrane Potential Distribution}
\label{membrane_potential_distribution}

To estimate the relative firing distributions of competing LIF neurons, it is useful to understand the distribution of their membrane potentials.
Assuming a firing event has yet to occur, the effect of a Poisson spike train on a neuron's membrane potential with exponential leakage may be viewed as a shot-noise process (\cite{PhysRevE.63.031902}).
A Poisson spike train from input $i$ is the summation of many spikes represented as delta functions:

\begin{equation}
N_i = \sum_k \delta_{T_{|ik|}}
\end{equation}

This stochastic process produces the following pre-firing membrane potential induced on neuron $j$ by the spike train from input $i$:

\begin{equation}
V_{ij}(t) = \int f_{ij}(t)N(dt) = \sum_k f_{ij}(t-T_k),
\end{equation}

where $f_{ij}(t) = w_{ij}e^{-t/\tau_{mem}}$.
The Laplace transform of this shot-noise process is:

\begin{equation}
\mathcal{L}(\theta) = E[e^{-\theta V_{ij}(t)}] = e^{g(\theta)}
\end{equation}

where $g(\theta) = \lambda_i\int_{0}^{t}(e^{-\theta f_{ij}(v)}-1)dv$.

\paragraph{Mean Pre-firing Membrane Potential}
\label{subsub_mean}

The $1^{st}$ moment, which is the mean pre-firing potential caused by input channel $i$, is given by:

\begin{align}
    E[V_{ij}(t)] &= -\Big[\frac{d\mathcal{L}(\theta)}{d\theta}\Big]_{\theta=0} = -\bigg[\frac{de^{g(\theta)}}{d\theta}\bigg]_{\theta=0}
    \nonumber\\
    &= -\Big[e^{g(\theta)}\Big]_{\theta=0}\Big[\frac{dg(\theta)}{d\theta}\Big]_{\theta=0}
    \nonumber\\
    &= -\lambda_i\bigg[\int_0^t(-f_{ij}(v)e^{-\theta f_{ij}(v)})dv\bigg]_{\theta=0}
    \nonumber\\
    &= \lambda_i\int_0^t f_{ij}(v)dv = \lambda_i w_{ij} \tau_{mem} (1-e^{-t/\tau_{mem}})
\end{align}

For all inputs, represented as the rate vector $\vec{\lambda}$, the mean combined pre-firing potential of neuron $j$ is:

\begin{align}
    E[V_j(t)] &= \tau_{mem}\sum_i\lambda_{i}w_{ij}(1-e^{-t/\tau_{mem}})
    \nonumber\\
    &= \tau_{mem}(\vec{w_j}\bullet\vec{\lambda})(1-e^{-t/\tau_{mem}})
    \label{EQ_combined_shot_noise_mean}
\end{align}

In steady-state this converges to: $\tau_{mem}(\vec{w_j}\bullet\vec{\lambda})$, which is important for discussions later in Sections \ref{sub_STDP} and \ref{subsub_mebrane_decay_time_constant}.

\paragraph{Variance of Pre-firing Membrane Potential}
\label{subsub_variance}

Continuing to the second moment, we can calculate the variance of the pre-firing membrane potential that is induced on neuron $j$ by incoming spikes received from input $i$:

\begin{align}
  \nonumber
  Var(V_{ij}(t)) &= E[V_{ij}(t)^2] - E[V_{ij}(t)]^2\\
  \nonumber
  &= \Big[\frac{d^2\mathcal{L}(\theta)}{d\theta^2}\Big]_{\theta=0} - E[V_{ij}(t)]^2\\
  \nonumber
  &= [e^{g(\theta)}(g'(\theta)^2+g''(\theta))]_{\theta=0} - E[V_{ij}(t)]^2\\
  \nonumber
  &= E[V_{ij}(t)]^2 + \lambda_{i}\int_0^t f_{ij}(v)^2 dv - E[V_{ij}(t)]^2\\
  &= \frac{1}{2}\lambda_i\tau_{mem} w_{ij}^2  (1-e^{-2t\tau_{mem}})
\end{align}

The combined variance of the potential induced by all inputs is:

\begin{align}
    \nonumber
    Var(V_j(t)) &= \frac{1}{2}\tau_{mem}\sum_i\lambda_{i}w_{ij}^{2}(1-e^{-2t\tau_{mem}})\\
    &= \frac{1}{2}\tau_{mem}(\vec{\lambda}\bullet\vec{w_j}^{\circ 2})(1-e^{-2t\tau_{mem}})
    \label{eq_final_variance}
\end{align}

where $\vec{w_j}^{\circ 2}$ represents the Hadamard square of the weight vector.
This equation is important for discussions later in Section \ref{subsub_Modified_STDP}.

\subsection{Methodology}

We simulated single layer CFNs on the completely disjoint MNIST dataset (\cite{lecun1998mnist}) on network sizes of 400, 900, 1600, 2500. 3600, 4900, and 6400 excitatory neurons.

\subsubsection{STDP Learning}
\label{sub_STDP}

The MNIST dataset is a magnitude insensitive dataset, meaning that increasing or decreasing the intensity of a sample does not alter its class and that angular distance is more important than Euclidean distance.
As given in (\ref{EQ_combined_shot_noise_mean}), the mean pre-firing potential of a spiking neuron is proportional to the L2-norm of its weight vector and also to the L2-norm of the input rate vector.
Although a larger mean pre-firing potential does not always correspond to a larger firing rate due to differing variances caused by the Hadamard square of the weight vector as shown in (\ref{eq_final_variance}), the correlation between $E[V]$ and the firing rate sufficiently holds for the MNIST dataset with inputs of large enough dimensions and fairly comparable input sparsity between samples.

As such, for a given input and assuming equal weight vector magnitudes, the neuron that is angularly closest to the input will be more likely to fire, allowing for unsupervised Hebbian learning by training neurons on correlated inputs. 
Therefore, we L2-normalize each neuron's weight vector.
Weight normalization has recently been shown to occur in biology (\cite{Boustani2018weightbalancing}) and may still be considered a localized function, as the processing can occur at the post-synaptic neuron to which all the weights in a given weight vector are directly connected.
The input rate vectors are also L2-normalized.

\paragraph{Stabilizing STDP}
\label{subsub_Modified_STDP}

STDP's Hebbian learning rule involves potentiation or depression of a synaptic weight based on the timing of pre- and post-synaptic firing events.
As the input information in our system is encoded only in the spike rate, we can employ the simple one-sided version of STDP, evaluated at the post-synaptic firing event:
\begin{equation} \label{eq_one_sided_STDP}
\Delta w = \alpha (pre - offset)
\end{equation}
where $\alpha$ is the learning rate (set to 0.01), $pre$ is a trace of pre-synaptic firing events, and $offset$ is the value to which the pre-synaptic traces are compared, determining potentiation or depression.

The $pre$ trace follows a similar distribution as the membrane potential (see Section \ref{membrane_potential_distribution}), only with a different time constant and without being weighted by the synapse, and so its expected value is also proportional to the input spike rate (e.g. $E[pre_i] = \lambda_i \tau_{pre}$).
Correlated potentiations in the direction of $\vec{pre}$ therefore provide Hebbian learning by angularly migrating $\vec{w}$ toward the angle of the input vector $\vec{\lambda}$.
Anti-Hebbian depression reduces weights from uncorrelated inputs and is provided by subtracting the $offset$ term for one-sided STPD rather than performing additional weight processing at pre-synaptic firing events.

\begin{figure}
\begin{center}
\includegraphics[width=176mm]{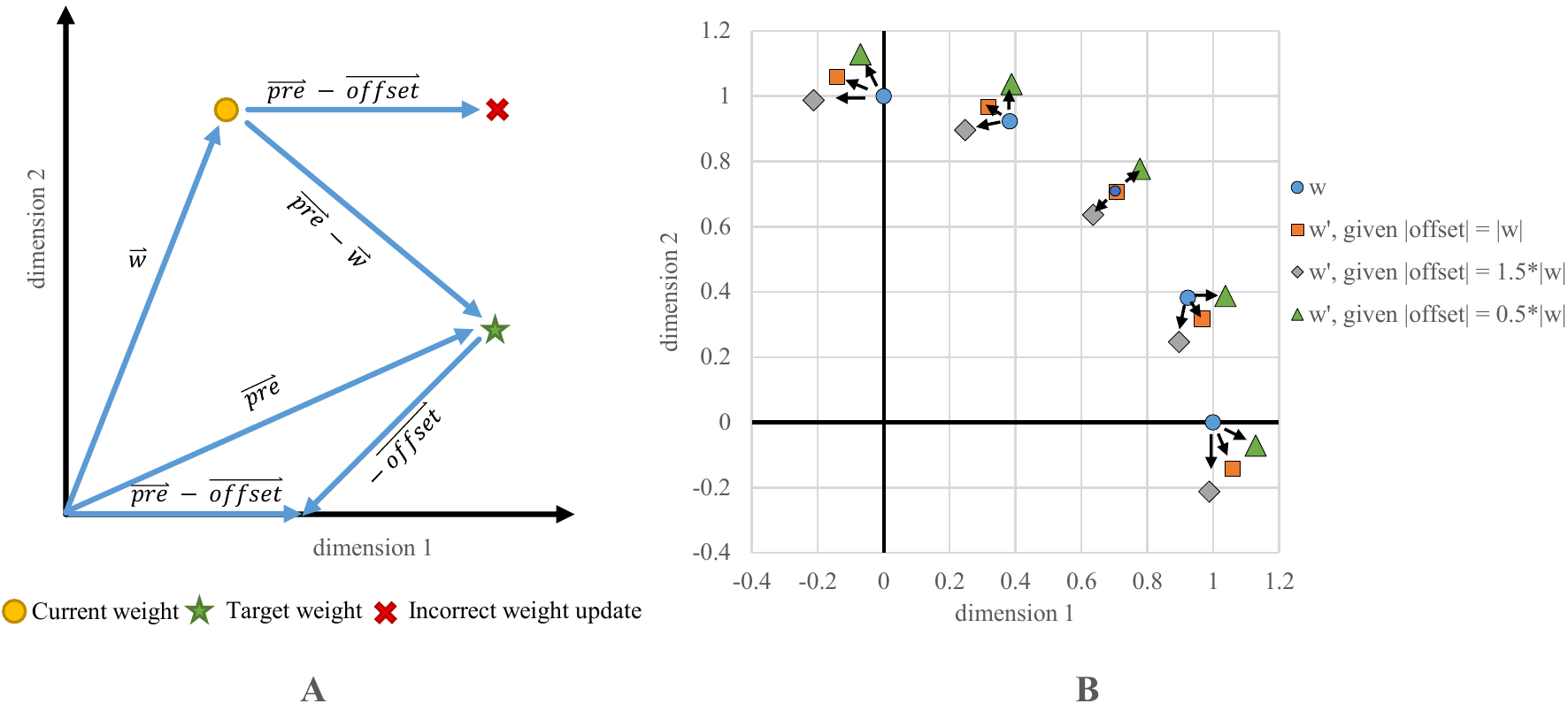}
\end{center}
\caption{Instability of one-sided STDP. \textbf{(A)} Example vectors showing how a static offset does not result in a correct weight change. The goal is to migrate $\vec{w}$ toward the $\vec{pre}$ trace, which is proportional to the input $\vec{\lambda}$. The $\vec{offset}$ vector that is subtracted from $\vec{pre}$ must be dynamically tied in each dimension to $\vec{w}$, rather than being the same in every dimension. \textbf{(B)} Weight change results for various starting positions where the target vector is equal to the current weight vector, which would ideally result in no weight change. With a static offset in each dimension, even scaled to the appropriate magnitude, the weight vectors do not stabilize on the target and instead migrate toward the axes, creating binarized weights when capped at zero.}
\label{fig:oja}
\end{figure}

Typically, $offset$ is a constant value identical across all dimensions and can be thought of as a scaled ones vector, applying uniform anti-Hebbian depression.
Such uniform depression does not, however, create a weight change in exactly the direction desired (see Figure \ref{fig:oja}), and causes instability in the STDP learning rule.
This instability is usually controlled by adaptive thresholding and weight capping via exponential weight-dependence.

However, our CFNs with rapid one-shot dopaminergic learning of novel inputs cannot use such gradual approaches to stabilize.
We provide the required stability to this STDP learning rule by correcting the direction of the weight change.
Rather than a constant offset, we dynamically tie $offset$ to the current weight value, which is an adaptation based on Oja's rule (\cite{Oja1982}).
To place $pre$ and the weight on the same scale, we scale the pre-synaptic trace by the inverse of its decay rate $\tau_{pre}$, changing (\ref{eq_one_sided_STDP}) to:

\begin{equation}
\Delta w = \alpha \Big(\frac{pre}{\tau_{pre}} - w\Big)
\end{equation}

This corrected weight change allows our CFNs to rapidly and accurately capture information from novel input during dopaminergic learning and otherwise gradually stabilize on the center of the cluster of input samples for which it has fired.

\subsubsection{Timing and Time Constants}

As our evaluations and simulations are purely event-driven, the concept of discrete computational time steps is not applicable. Timing parameters are therefore purely relative, and so without loss of generality, we L2 normalize the magnitude of the input rate vectors to 1 spike per time unit.

\paragraph{Membrane Decay Time Constant}
\label{subsub_mebrane_decay_time_constant}

According to Equation (\ref{EQ_combined_shot_noise_mean}), the expected value of the membrane potential saturates in time according to $(1-e^{-t/\tau})$. A smaller $\tau$ results in a faster convergence to the steady state, or, equivalently, fewer input spikes to converge. E.g., in five time constants, the expected potential reaches over 99\% of is steady-state value. However, using (\ref{eq_final_variance}), the steady state standard deviation of the potential in proportion to the mean decreases as the decay rate increases:

\begin{equation}
    \frac{\sqrt{Var(V)}}{E[V]} \propto \frac{1}{\sqrt{\tau}}
\end{equation}

Thus, a larger membrane decay constant is better for proper discrimination between two differing inputs, but increases the number of computations. For the L2-normalized MNIST dataset with 784 input dimensions, the angular distances between samples of differing classes are close enough to require at least 10 to 15 normalized time units for $\tau_{mem}$ in order to successfully establish a firing threshold that can discriminate between classes.

\paragraph{Time to Recognize}

A $\tau_{mem}$ of 15 still produces enough variance according to (\ref{eq_final_variance}) that two to three time constants is on average sufficient time for the potential to rise above its steady-state mean.
Following the setup in \cite{Diehl_MNIST}, we identify successful recognition of an input sample after registering five output sikes.
Therefore, a total of 150 to 225 time units was generally sufficient to produce five sequential firing events in a reference vector neuron with a center close to the input.

In our simulations, we found little accuracy change by adjusting this hyperparameter within this range as long as the threshold voltage was appropriately tuned, so we fixed the time to recognize at 200 normalized time units for each simulation.
We tuned the dopaminergic neuron to fire after those 200 time units unless it has been otherwise inhibited as discussed in Section \ref{Section_Dop}.
We also set $\tau_{pre}$ to this value to capture as much of the input train as possible because of the rapid one-shot dopaminergic learning of novel samples.

\subsubsection{Determining $v_{th}$ Without Adaptive Thresholding}

Short-term adaptive thresholding interferes with controlled forgetting.
Long-term adaptive thresholding may still be used with controlled forgetting if properly tuned, but enhancing plasticity and stimulated firing of infrequently-firing neurons is itself a form of deliberate, controlled homeostasis.
Therefore, for a more accurate evaluation of our CFNs, we have removed traditional adaptive thresholding completely. 
With normalized weight vectors and input vectors, the larger the ratio $v_{th}:E[V(t)]$ the closer the input rate vector must be angularly to the weight vector to produce a given firing probability.
Determining the proper $v_{th}$ without dynamic adaptation, therefore, depends on the tightness of the clustering in the dataset.
With this context, we included $v_{th}$ in our hyper-parameter search.

\subsubsection{Hyper-parameter Sweep}

SNNs are known to be highly sensitive to hyper-parameters, especially during unsupervised learning without error signals to provide dynamic corrections.
We perform a small search in the hyper-parameter space, adjusting $v_{th}$ and the number of training epochs.
Results from this search are shown in Table \ref{table:parameter_sweep}, with hyperparameters resulting in the best accuracy highlighted for each size.
Good machine learning practice requires that we choose the system parameters based only on the training set, so only training set accuracy results are shown here.
Testing accuracy results are discussed later in the Results section.

\begin{table}
\caption{Training accuracy results of hyper-parameter sweep for each network size across both $v_{th}$ and number of training epochs per task. Highlighted cells are best configuration for each size.}
\label{table:parameter_sweep}
\begin{center}
\begin{tabular}{|c|c|c|c|c|c|} 
 \cline{3-6}
 \multicolumn{2}{c|}{} & \multicolumn{4}{c|}{\# of Training Epochs per Task} \\
 \hline
 Neurons & $v_{th}$ & 1 & 5 & 10 & 20 \\ [0.5ex] 
 \hline
 \multirow{4}{*}{400} & 13.5 & \cellcolor{lightgray} 87.53\% & 83.74\% & 79.03\% & 73.94\% \\
 \cline{2-6}
  & 13.75 & 87.43\% & 83.97\% & 78.20\% & 75.06\% \\
 \cline{2-6}
  & 14 & 86.47\% & 82.32\% & 77.03\% & 75.09\% \\
 \cline{2-6}
  & 14.25 & 85.30\% & 83.46\% & 75.70\% & 74.94\% \\
 \hline
 \multirow{4}{*}{900}  & 13.5 & 89.74\% & 90.97\% & 89.25\% & 85.15\% \\
 \cline{2-6}
  & 13.75 & 88.87\% & 90.87\% & 89.71\% & 86.11\% \\
 \cline{2-6}
  & 14 & 87.68\% & \cellcolor{lightgray} 91.35\% & 89.46\% & 84.62\% \\
 \cline{2-6}
  & 14.25 & 86.54\% & 91.26\% & 89.82\% & 84.02\% \\
 \hline
 \multirow{4}{*}{1600}  & 13.5 & 91.48\% & 92.29\% & 92.22\% & 91.15\% \\
 \cline{2-6}
  & 13.75 & 90.99\% & 92.69\% & 92.33\% & 91.32\% \\
 \cline{2-6}
  & 14 & 89.92\% & \cellcolor{lightgray} 93.12\% & 92.13\% & 91.21\% \\
 \cline{2-6}
  & 14.25 & 88.31\% & 93.02\% & 92.73\% & 91.67\% \\
 \hline
 \multirow{4}{*}{2500}  & 13.5 & 93.14\% & 93.50\% & 93.37\% & 92.99\% \\
 \cline{2-6}
  & 13.75 & 92.81\% & 93.57\% & 93.93\% & 93.10\% \\
 \cline{2-6}
  & 14 & 91.62\% & 93.35\% & \cellcolor{lightgray} 94.20\% & 93.49\% \\
 \cline{2-6}
  & 14.25 & 89.92\% & 92.95\% & 93.86\% & 93.30\% \\ 
 \hline
 \multirow{4}{*}{3600} & 13.5 & 93.76\% & 94.07\% & 93.95\% & 93.69\% \\
 \cline{2-6}
  & 13.75 & 93.76\% & 93.98\% & 94.42\% & 94.44\% \\
 \cline{2-6}
  & 14 & 93.19\% & 93.91\% & 94.41\% & \cellcolor{lightgray} 94.49\% \\
 \cline{2-6}
  & 14.25 & 92.19\% & 93.15\% & 94.21\% & 94.22\% \\
 \hline
  \multirow{4}{*}{4900} & 13.5 & 94.49\% & 94.81\% & 94.66\% & 94.64\% \\
 \cline{2-6}
  & 13.75 & 94.91\% & 94.92\% & 94.78\% & 95.04\% \\
 \cline{2-6}
  & 14 & 94.59\% & 94.74\% & 94.97\% & \cellcolor{lightgray} 95.17\% \\
 \cline{2-6}
  & 14.25 & 93.52\% & 93.79\% & 94.51\% & 95.16\% \\
 \hline
 \multirow{4}{*}{6400} & 13.5 & 95.40\% & 95.23\% & 95.25\% & 95.21\% \\
 \cline{2-6}
  & 13.75 & 95.32\% & 95.57\% & 95.33\% & 95.72\% \\
 \cline{2-6}
  & 14 & 95.37\% & 95.50\% & 95.39\% & 95.76\% \\
 \cline{2-6}
  & 14.25 & 94.67\% & 94.86\% & 95.11\% & \cellcolor{lightgray} 95.79\% \\
 \hline
\end{tabular}
\end{center}
\end{table}

\paragraph{Neuron Firing Thresholds, $v_{th}$}

Based on the discussion above, $v_{th}$ should be close to but slightly less than $\tau_{mem}$ in voltage units, which is set to 15 time units.
For MNIST, we initially found that if $v_{th}$ is much less than 13.5, a neuron may too likely fire for samples from other classes, while if $v_{th}$ is much higher than 14.25, a neuron may not fire for very close samples, even different stochastic instances of the same sample.
We therefore tested each setup with four different threshold values in this range: 13.5, 13.75, 14.0, and 14.25.
Smaller networks require each individual neuron to capture a larger subset of input samples, generally requiring slightly lower thresholds than those in larger networks.

\paragraph{Number of Training Epochs}

Larger networks can capture representations that are less common but still useful.
As such, for larger networks more epochs within a class are required before proceeding to subsequent tasks in order to refine the less common representations.
For smaller networks, on the other hand, more epochs may reinforce less useful outliers, making it more difficult to make room for subsequent tasks.

\subsubsection{Comparison of $E[V(t)]$ at $v_{th}$ with K-Means Clustering Angular Error.} 

\begin{figure}
\begin{center}
\includegraphics[width=85mm]{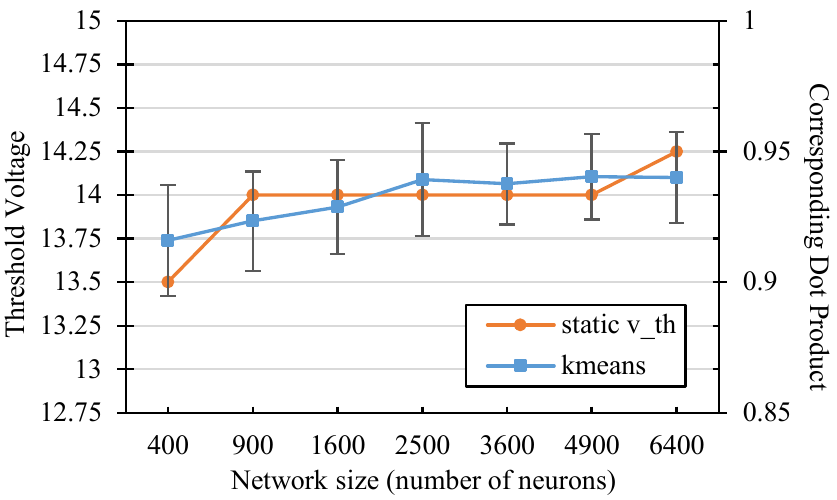}
\end{center}
\caption{Comparison of the static $v_{th}$ selected in the hyperparameter sweep with the corresponding dot product of the nearest training error in a kmeans network of the same size. The kmeans error bars represent two standard deviations over 100 trials each.}
\label{fig_kmeans}
\end{figure}

We can compare the $v_{th}$ values selected in the hyper-parameter search with the mean angular distance to a neuron's weight vector that would on average result in a membrane potential equal to that threshold.
Performing a simple k-means clustering on the L2-normalized MNIST dataset yields information on the relative desired scope of each reference vector, depending on the number of reference vector neurons.
Figure \ref{fig_kmeans} shows the dot product associated with the angular distance of the closest training sample / reference vector pair from differing classes for each network size after k-means clustering.
The figure also shows the average membrane potential of a spiking neuron corresponding to these angles.
For SNNs, neurons that are able to fire for samples that are further away than these angles are thus more likely to fire for samples of the wrong class.
As the number of reference vector neurons increases, the portion of the input space per neuron decreases, improving accuracy by allowing each individual neuron to be more restrictive in is angular scope, which is relatively similar to those associated with the $v_{th}$ values selected in the hyper-parameter sweep.

\subsubsection{Simulation Setup}

Using exponential kernels, we treat spikes as inducing instantaneous voltage potentiations in the respective post-synaptic neuron membranes with exponential decay.
As such, neurons only fire upon receiving an incoming spike and will not fire between incoming spikes, with the exception of the dopaminergic neurons which are handled separately.
This allows us to emulate the networks using purely event-driven computation rather than breaking time into discrete time steps and updating neurons states at each time step.
Because we encoded input spike trains as Poisson point processes, the time between spikes is an exponential random variable with $\lambda_i = input_i$.
Therefore, rather than incrementing time in fixed intervals, we calculate the time until the next input spike arrival and decay all the traces and membrane potentials according to that time interval before processing that input spike.

The dopaminergic neurons are an exception, as they fire in the \textit{absence} of input spikes.
However, with the membrane potential growth rate towards the elevated resting potential, we can calculate how long it would take to reach the dopaminergic neuron's firing threshold in the absence of input spikes.
Therefore, before processing an input spike, we first check to see if the dopaminergic neuron would have fired earlier, in which case, it is processed at its respective time interval first.

\paragraph{Training}

Samples from the current task/digit were presented one-by-one to the network before proceeding to the next task/digit.
Once a task change occurs, samples from previous tasks are never revisited.
The network receives no information about when a task change occurs.
For the current sample, input neurons fire at the sample rate until the system registers at least five output spikes, which is generally enough to confidently identify the input in view of the stochasticity.
During that time, we do not increase the input firing rate, since the dopaminergic neuron takes care of stimulating neurons in the absence of a good match.

\paragraph{Testing}

To show effective lifelong learning over time, we evaluate the testing accuracy results of all previously seen tasks at each task change.
The system is designed for online learning, updating weights via STDP whenever a sample is presented.
In deployment, therefore, learning and inference occur simultaneously.
Thus, to demonstrate performance on the testing set in our results, we must pause online learning and freeze network parameters during each testing set evaluation.

Training is performed entirely without supervision and without knowledge of a task change, and so the final network outputs must be assigned class labels before testing.
Proper machine learning requires that we assign class labels only using the training set.
Therefore, we also freeze network parameters during label assignment to avoid replay from previously-seen samples in the training set.
Label assignment is performed by inference on the training set, assigning each output neuron the class for which it dominated or fired most frequently.
With frozen parameters, dopaminergic adaptation also does not occur during assignment and testing.

Instead, a poorly-recognized input is assigned the closest class by continuing to increase input firing rates until a sufficient response is recorded.

\section{Results}

Using the hyper-parameter selection for each network size discussed above, we simulate online dopaminergic learning on the completely disjoint MNIST dataset, compared with an identical offline setup allowed to access the MNIST dataset in randomized order allowing for class interleaving.

\subsection{Final Results and the Expected ``Sequential Penalty"}

\begin{figure}
\begin{center}
\includegraphics[width=85mm]{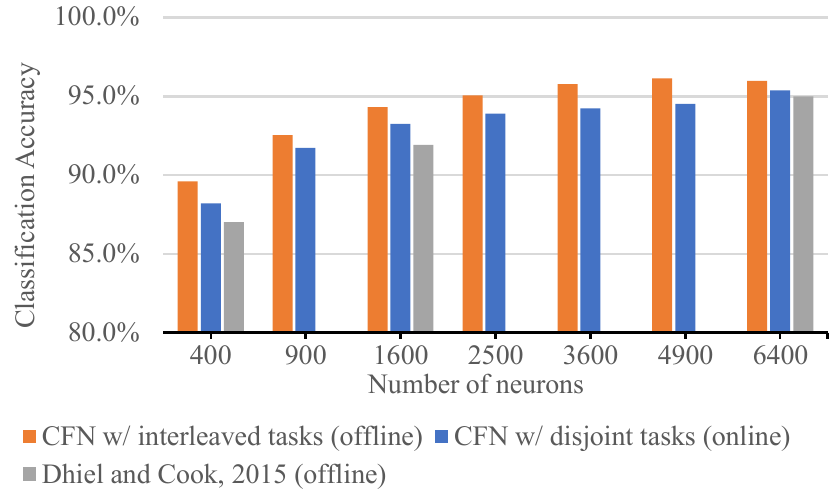}
\end{center}
\caption{Final classification accuracy of our CFNs at various sizes. The offline CFN has access to interleaved data. The online CFN shows comparable accuracy while successfully achieving lifelong learning with the tasks completely disjoint and sequential.}
\label{fig_final_total_comparison}
\end{figure}

Figure \ref{fig_final_total_comparison} shows the final results of our CFNs after learning all ten disjoint tasks sequentially without any data reinforcement from previous classes and compares these results to the accuracy the system would have achieved had it been allowed the clairvoyance of temporally distributing classes uniform-randomly throughout the training process.
We see that our lifelong SNNs perform on par with the interleaved systems, averaging 1.17\% accuracy reduction across all sizes.
This penalty is expected due to sequentializing the tasks, but is acceptable given the system’s avoidance of catastrophic failure without the data reinforcement provided by interleaved input classes.
In fact, even with this penalty, the 6400 neuron SNN achieves a respectable \textbf{95.36\%} test accuracy after lifelong learning, which we believe is the best unsupervised accuracy ever achieved by a fixed-size, single-layer SNN on a completely disjoint MNIST dataset.
Our CFNs even outperform \cite{Diehl_MNIST} in all cases for which they provide results, even though that work is offline with an interleaved dataset.

\subsection{Graceful Degradation Instead of Catastrophic Forgetting}

\begin{figure}
\begin{center}
\includegraphics[width=176mm]{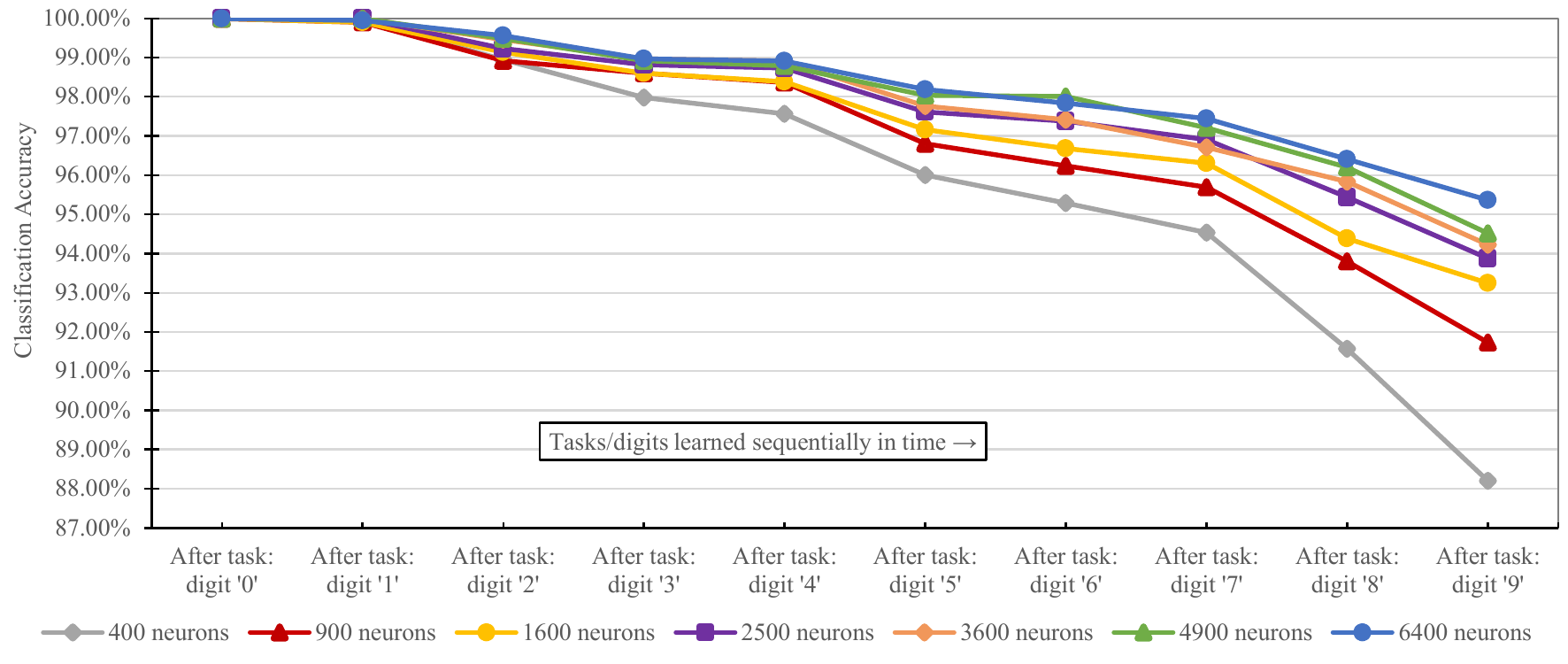}
\end{center}
\caption{Graceful accuracy degradation at each stage of the learning process (i.e. after each new task/digit) instead of catastrophic failure. Accuracy is for all previous tasks, up to and including the current task.}
\label{fig_acc_deg_test}
\end{figure}

The true success of a lifelong learning system is shown not just by the final accuracy, but also by its performance throughout the training process.
Figure \ref{fig_acc_deg_test} shows graceful degradation of total accuracy as each network broadens its capability while adding more training tasks to its limited resources.
They key is that there are no sharp drops in accuracy, which would occur if the system began failing at older tasks when a new task is learned.

\begin{figure}
\begin{center}
\includegraphics[width=85mm]{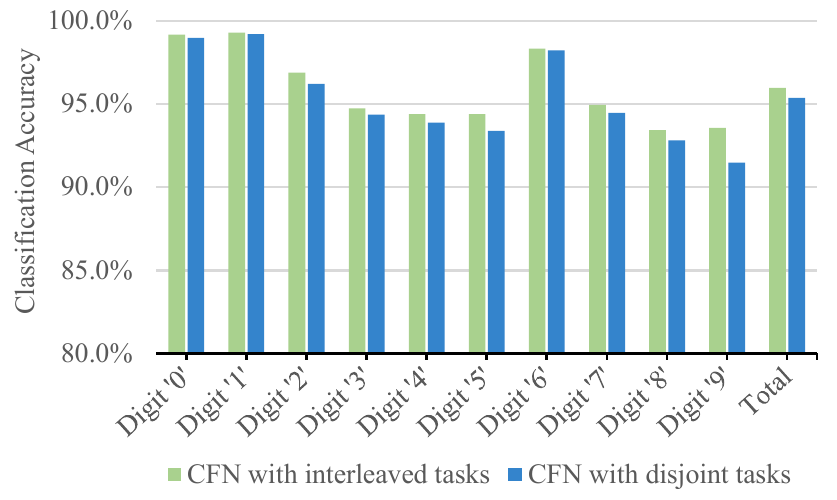}
\end{center}
\caption{Final per-digit accuracy after disjoint tasks of sequentially presented digit classes, compared to final per-digit accuracy when digits are interleaved (size 6400).}
\label{fig_per_digit_final}
\end{figure}

Further, the adaptation is distributed between tasks.
Figure \ref{fig_per_digit_final} shows the final accuracy of each individual task by the end of the training process for a CFN of 6400 neurons.
Notice that while the system expectedly performs better for some tasks rather than others, there is no single task for which the system fails; i.e., the sequential penalty is spread between tasks.
In fact, the lifelong system performs best at the same tasks (digits ‘0,’ ‘1,’ and ‘6’) and worst at the same tasks (digits ‘8’ and ‘9’) that the offline/non-lifelong system does.

Per-task test accuracy over time and per-task false positives over time are provided in the supplementary material, demonstrating that individual tasks also experience a graceful adaptation over time.

\section{Discussion}

\begin{figure}
\begin{center}
\includegraphics[width=85mm]{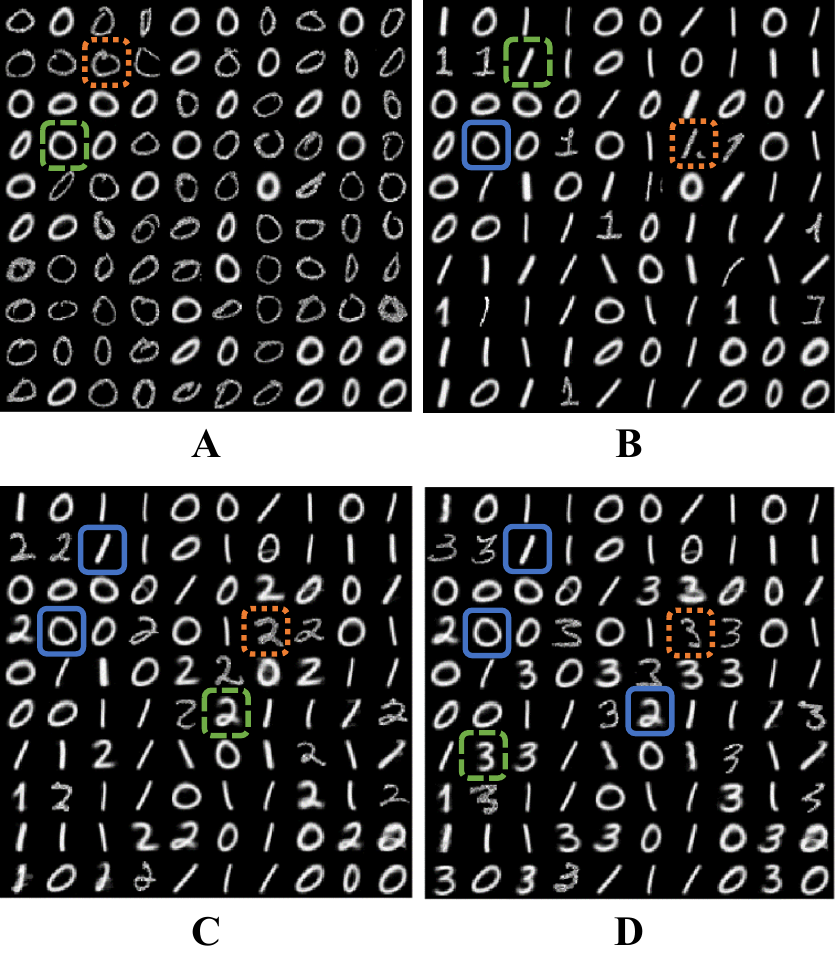}
\end{center}
\caption{Grid view of weight vectors of 100 reference neurons over time, trained with dopaminergic stimulation of STDP on four sequential tasks: the MNIST digits ‘0’ through ‘3.’ Digits highlighted in dashed green are examples of successfully learned generalized representations. Digits highlighted in dotted orange are examples of outlier representations. Digits highlighted in solid blue are examples of representations preserved from previous tasks. \textbf{(A)} Weights after learning digit `0.' \textbf{(B)} Weights after subsequently learning digit `1.' \textbf{(C)} Weights after subsequently learning digit `2.' \textbf{(D)} Weights after subsequently learning digit `3.'}
\label{fig_100_visual}
\end{figure}

In these fully-connected one-layer SNNs, each neuron’s weight vector can be viewed as a reference vector that captures a specific input representation, ideally successfully generalized.
As such, we may qualitatively observe the success of dopaminergic learning over time by viewing these representations.
For a better visual demonstration, we show the weights of a smaller 100-neuron CFN in Figure \ref{fig_100_visual}, arranged in a 10x10 grid, after each of four different learning tasks–digits ‘0’ through ‘3’ from the MNIST dataset.
The digit representations that are more smooth and solid are generalized representations, while the representations that appear less defined are outlier representations.

Notice that the digit representations that are preserved from one task to another are the useful generalizations rather than the outliers, which on the other hand are the first to be overwritten when space for a new task is required.
In this memory-constrained example of only 100 neurons, by the time the system gets to digit 3, it runs out of old outliers to replace.
Therefore, in order to make space for the new information, it must unavoidably replace some useful generalized representations from previously learned tasks.
However, whether outlier or generalized representations are replaced, notice that the representations that are preserved from previous tasks experience very little and infrequent corruption during later learning stages.
The dopamine signals are able to successfully replace old information with new information without interference because of the targeted localization provided by stimulating STDP.
This controlled forgetting allows the network to gracefully degrade its accuracy, especially for older tasks, in exchange for the ability to learn new tasks.
If training samples had been presented sequentially in this manner to a traditional SNN, the samples from later tasks would have significantly corrupted the previously learned representations, making the network entirely useless at classification.
Such failure is successfully avoided by our method of controlled forgetting.

\subsection{Future Work}

We expect that a deeper network will improve accuracy beyond that of these results and allow for learning of more complicated datasets.
As mentioned earlier, in a deeper network, it may be that only the last few layers would require lifelong learning, performing a readout from a liquid state machine or a fixed feed forward network sufficiently pre-trained on low-level representations.
We also plan to evaluate this method on time-encoded signals.
Further, we hope to explore other dopaminergic weight adjustment policies that have a higher time-dependence or weight policies with habituation, such as in \cite{panda2018asp}, in order to allow for operation in an environment of changing tasks, and not just temporally separated tasks.

\subsection{Conclusion}

We presented a biologically-inspired dopaminergic modulation of synaptic plasticity to exploit STDP locality.
Trained stimulation during the presentation of novel inputs allows the system to quickly perform isolated adaptation to new information while preserving useful information from previous tasks.
This method of controlled forgetting successfully achieves lifelong learning.
Our Controlled Forgetting Networks show only a slight reduction in accuracy when given the worst possible class ordering, i.e. completely sequential without any data reinforcement between classes, while successfully avoiding catastrophic forgetting.

\section*{Conflict of Interest Statement}

The authors declare that the research was conducted in the absence of any commercial or financial relationships that could be construed as a potential conflict of interest.

\section*{Author Contributions}

JA wrote the paper and performed the simulations.
All authors helped with developing the concepts, conceiving the experiments, and writing the paper.
The authors would also like to thank Professor Jonathon Peterson and Dr. Gerard (Rod) Rinkus of Purdue University for helpful discussions.

\section*{Funding}

This work was supported in part by C-BRIC, a JUMP center sponsored by the Semiconductor Research Corporation and DARPA, and by the National Science Foundation, Intel Corporation, and the Vannevar Bush Fellowship.

\section*{Data Availability Statement}

The MNIST dataset used in this study can be found at \href{http://yann.lecun.com/exdb/mnist}{http://yann.lecun.com/exdb/mnist}.

\bibliographystyle{frontiersinSCNS_ENG_HUMS} 
\bibliography{ControlledForgetting}

\end{document}

\onecolumn
\firstpage{1}

\title[Supplementary Material]{{\helveticaitalic{Supplementary Material}}}

\maketitle

\section{Alternative Derivation of Expected value of Pre-firing Membrane Potential}

Let $V_j(t)$ be the random variable representing the potential of neuron $j$ at time $t$.
First, we consider the effect of the most recent incoming spike $k$ which is received from input $i$ at time $t_{ik}$.
If at the time immediately before the spike was received, $t^-_{ik}$, the voltage of neuron $j$ over the resting potential $v_{rest}$ is given by $v_j(t^-_{ik})$, then, assuming neuron $j$ has not since fired, the potential of neuron $j$ at time $t$ is given by:
\begin{equation} \label{potential_most_recent_spike}
v_j(t) = \big(v_j(t^-_{ik}) + w_{ij}u(t-t_{ik})\big)e^{-(t-t_{ik})/\tau_{mem}} + v_{rest}
\end{equation}
where $u(t)$ is the unit step function.
If we assume that the input spike has already occurred, i.e. $t > t_{ik}$, then the unit step function may be removed.
Provided that the decay rate $\tau_{mem}$ remains constant, we see that the value of the potential immediately preceding the spike, $v_j(t^-_{ik})$, and the weighted potential increase induced by the spike, $w_{ij}$, may be linearly separated.
Without loss of generality, this linear separability may be extended to the residual potential increases induced by all spikes received by a neuron since its last firing event and refractory period, given by:

\begin{equation} \label{potential_all_spikes}
v_j(t) =  v_{rest} + \sum_i \sum_k w_{ij}e^{-(t-t_{ik})/\tau_{mem}}
\end{equation}

As discussed in the text, we assume that $v_{reset} = v_{rest}$.
In cases where this does not hold, the reset voltage is also linearly separable, and its residual effect may simply be added as $+(v_{reset}-v_{rest})e^{-(t-t_{last\_ref})/\tau_{mem}}$ where $t_{last\_ref}$ is the time since the last refractory period.

Since $t_{|ik|} = t - t_{ik}$ follows the distribution of gamma random variable $T_{|ik|} \sim gamma(\alpha=k;\beta=\lambda_i)$, let
\begin{equation} \label{function_on_random}
V_{|ik|} = u(T_{|ik|}) = e^{-T_{|ik|}/\tau_{mem}}
\end{equation}
be the random variable representing the unweighted portion of the potential increase from the $k^{th}$ most recent spike from $i$.
This allows us to rewrite (\ref{potential_all_spikes}) as a random variable:

\begin{equation} \label{sum_of_random}
V_j = v_{rest} + \sum_i \sum_k w_{ij}V_{|ik|}
\end{equation}

Because $T_{|ik|}$ is a gamma random variable, we know its CDF is
\begin{equation} \label{cdf_t}
F_{T_{|ik|}}(t) = \frac{\gamma(k,\lambda_{i}t)}{\Gamma(k)}
\end{equation}
where $\Gamma(s)$ and $\gamma(s,x)$ are the gamma function and the lower incomplete gamma function, respectively.

We can use (\ref{cdf_t}) and invert (\ref{function_on_random}) as (\ref{invert}) to solve for the transformation from $T_{|ik|}$ to $V_{|ik|}$ in (\ref{cdf}).

\begin{equation} \label{invert}
t_1 = u^{-1}(v) = -\tau_{mem} \ln v
\end{equation}

\begin{equation} \label{cdf}
F_{V_{|ik|}}(v) = 1 - F_{T_{|ik|}}(t_1) = \frac{\Gamma(k,-\lambda_i\tau_{mem}\ln v)}{\Gamma(k)}
\end{equation}
where $\Gamma(s,x)$ is the upper incomplete gamma function.

We take the derivative of the CDF in (\ref{cdf}) to give us the following PDF for $V_{|ik|}$:

\begin{equation} \label{pdf}
f_{V_{|ik|}}(v) = \frac{(\lambda_i\tau_{mem})^k v^{\lambda_i\tau_{mem}-1}(-\ln v)^{k-1}}{(k-1)!}
\end{equation}

We use the pdf of $V_{|ik|}$ to calculate its expected value in (\ref{expected_single}), and since expectation is a linear operator, we solve for the expectation of $V_{j}$ in (\ref{expected_all}) from (\ref{sum_of_random}) and (\ref{expected_single}).

\begin{equation} \label{expected_single}
E(V_{|ik|}) = \int_0^1 vf_{V_{|ik|}}(v)dv = \frac{(\lambda_i\tau_{mem})^k}{(1+\lambda_i\tau_{mem})^k}
\end{equation}

\begin{equation} \label{expected_all}
E(V_j) = v_{rest} + \sum_i \sum_k w_{ij}\frac{(\lambda_i\tau_{mem})^k}{(1+\lambda_i\tau_{mem})^k}
\end{equation}

If we take the sum of spikes to infinity to get the steady state, which is a reasonable approximation since only the most recent spikes have a significant impact on the potential, then the inner sum converges to:
\begin{align}
E(V_j) & \approx v_{rest} + \sum_i w_{ij}\lambda_i\tau_{mem} \nonumber \\
& \approx v_{rest} +  \tau_{mem} (\vec{w_{j}} \bullet \vec{\lambda})
\end{align}
which includes a simple scaled dot product of neuron $j$'s weight vector $w_j$ and the input rate vector $\lambda$, as you would find in a non-spiking neuron.

The appropriateness of this approximation is strengthened by the fact that in paper, the equation is used not to determine the precise spiking rate of an individual neuron but rather to compare relative spiking rates between competing neurons.

\section{CFN Lifelong Learning Results Broken Down by Task over Time}

\begin{figure}
\begin{center}
\includegraphics[width=176mm]{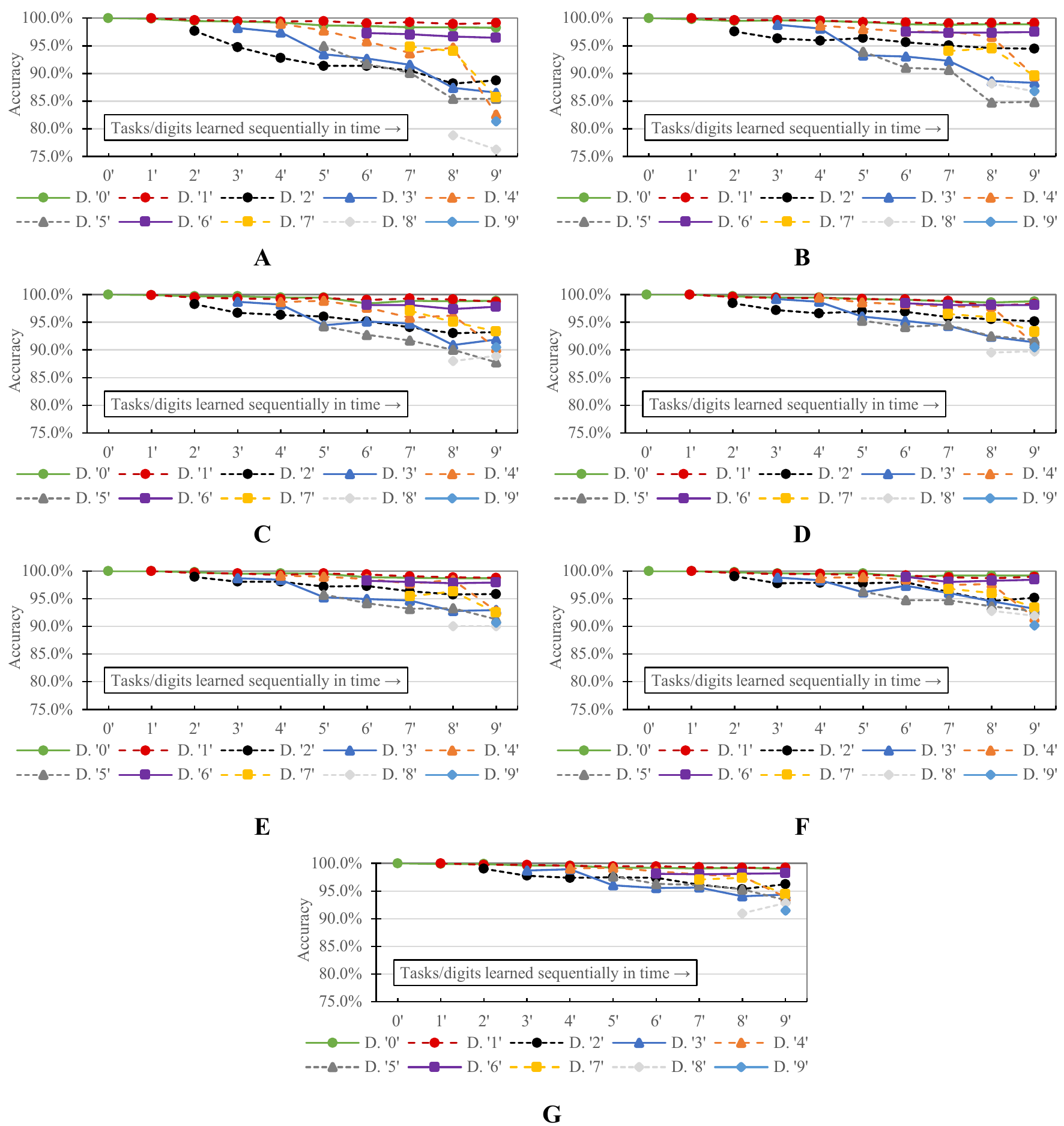}
\end{center}
\caption{Per-task/digit classification accuracy as new tasks/digits are added over time using CFNs with \textbf{(A)} 400 neurons, \textbf{(B)} 900 neurons, \textbf{(C)} 1600 neurons, \textbf{(D)} 2500 neurons, \textbf{(E)} 3600 neurons, \textbf{(F)} 4900 neurons, and \textbf{(G)} 6400 neurons.}
\label{fig_acc_per_digit}
\end{figure}

\begin{figure}
\begin{center}
\includegraphics[width=176mm]{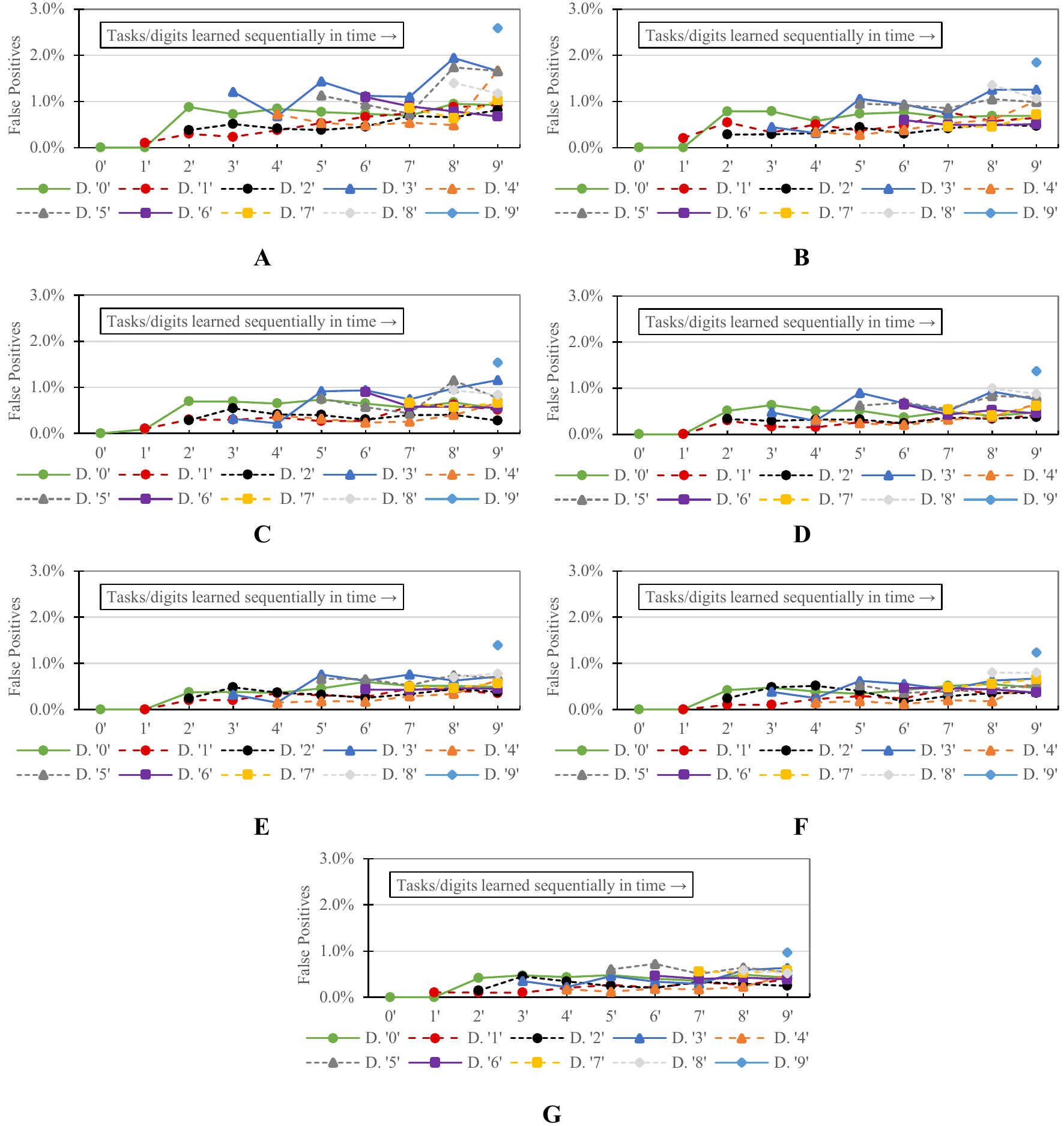}
\end{center}
\caption{Per task/digit false positives as new tasks/digits are added over time using CFNs with \textbf{(A)} 400 neurons, \textbf{(B)} 900 neurons, \textbf{(C)} 1600 neurons, \textbf{(D)} 2500 neurons, \textbf{(E)} 3600 neurons, \textbf{(F)} 4900 neurons, and \textbf{(G)} 6400 neurons.}
\label{fig_fals_pos_per_digit}
\end{figure}

Figure \ref{fig_acc_per_digit} shows the per-task test accuracy over time, and Figure \ref{fig_fals_pos_per_digit} shows the per-task false positives throughout the training process as new tasks are added, demonstrating graceful adaption to new tasks rather than catastrophic failure.

